\documentclass[a4paper]{article}
\usepackage{twocolceurws}

\usepackage{booktabs} 
\usepackage{bbm}
\usepackage{algorithm}
\usepackage{times}
\usepackage{algorithmic}
\usepackage{graphicx}
\usepackage{amsmath}
\usepackage{amsfonts}

\usepackage[tight,footnotesize]{subfigure}
\usepackage{url}
\usepackage{bm}
\usepackage{nicefrac}
\usepackage{subfigure}
\usepackage{caption}
\usepackage{enumitem}
\usepackage{multirow}
\usepackage{color}
\usepackage{titlesec}
\usepackage[normalem]{ulem}

\usepackage{cite}

\usepackage{array}
\newcolumntype{L}[1]{>{\raggedright\let\newline\\\arraybackslash\hspace{0pt}}m{#1}}
\newcolumntype{C}[1]{>{\centering\let\newline  \\\arraybackslash\hspace{0pt}}m{#1}}
\newcolumntype{R}[1]{>{\raggedleft\let\newline \\\arraybackslash\hspace{0pt}}m{#1}}

\usepackage{pdftexcmds}
\usepackage{catchfile}
\usepackage{ifluatex}
\usepackage{ifplatform}

\ifwindows
\usepackage{txfonts}
\usepackage{times}

\fi
\iflinux
\fi
\ifmacosx
\fi 

\usepackage{array,ragged2e}

\newcolumntype{P}[1]{>{\RaggedRight\hspace{0pt}}p{#1}}

\graphicspath{ {figures/} }

\newsavebox{\tempfig}

\def\bW{\textbf{W}}
\def\bX{\textbf{X}}

\def\bh{\textbf{h}}

\def\bz{\textbf{z}}

\title{Simplifying Architecture Search for Graph Neural Network}

\author{Huan Zhao$^1$, \quad Lanning Wei$^2$, \quad Quanming Yao$^3$ \\
	\{zhaohuan, weilanning, yaoquanming\}@4paradigm.com}

\institution{4Paradigm Inc. Shenzhen$^1$, Hong Kong$^3$.
\\
Institute of Computing Technology Chinese Academy of Sciences$^2$}

\begin{document}

\maketitle

\begin{abstract}
Recent years have witnessed the popularity of Graph Neural Networks (GNN) in various scenarios.  
To obtain optimal data-specific GNN architectures, researchers turn to neural architecture search (NAS) methods, 
which have made impressive progress in discovering effective architectures in convolutional neural networks. 
Two preliminary works, 
GraphNAS and Auto-GNN, have made first attempt to apply NAS methods to GNN. 
Despite the promising results, 
there are several drawbacks in expressive capability and search efficiency of GraphNAS and Auto-GNN due to the designed search space. 
To overcome these drawbacks, 
we propose the SNAG framework (Simplified Neural Architecture search for Graph neural networks), 
consisting of a novel search space and a reinforcement learning based search algorithm. 
Extensive experiments on real-world datasets demonstrate the effectiveness of the SNAG framework compared to human-designed GNNs and NAS methods, including GraphNAS and Auto-GNN.\footnote{	Correspondence is to Quanming Yao;
	and the work was done when Lanning Wei was an intern in 4Paradigm.}
\end{abstract}

\begin{figure*}[ht]
	\centering
	\includegraphics[width=0.8\textwidth]{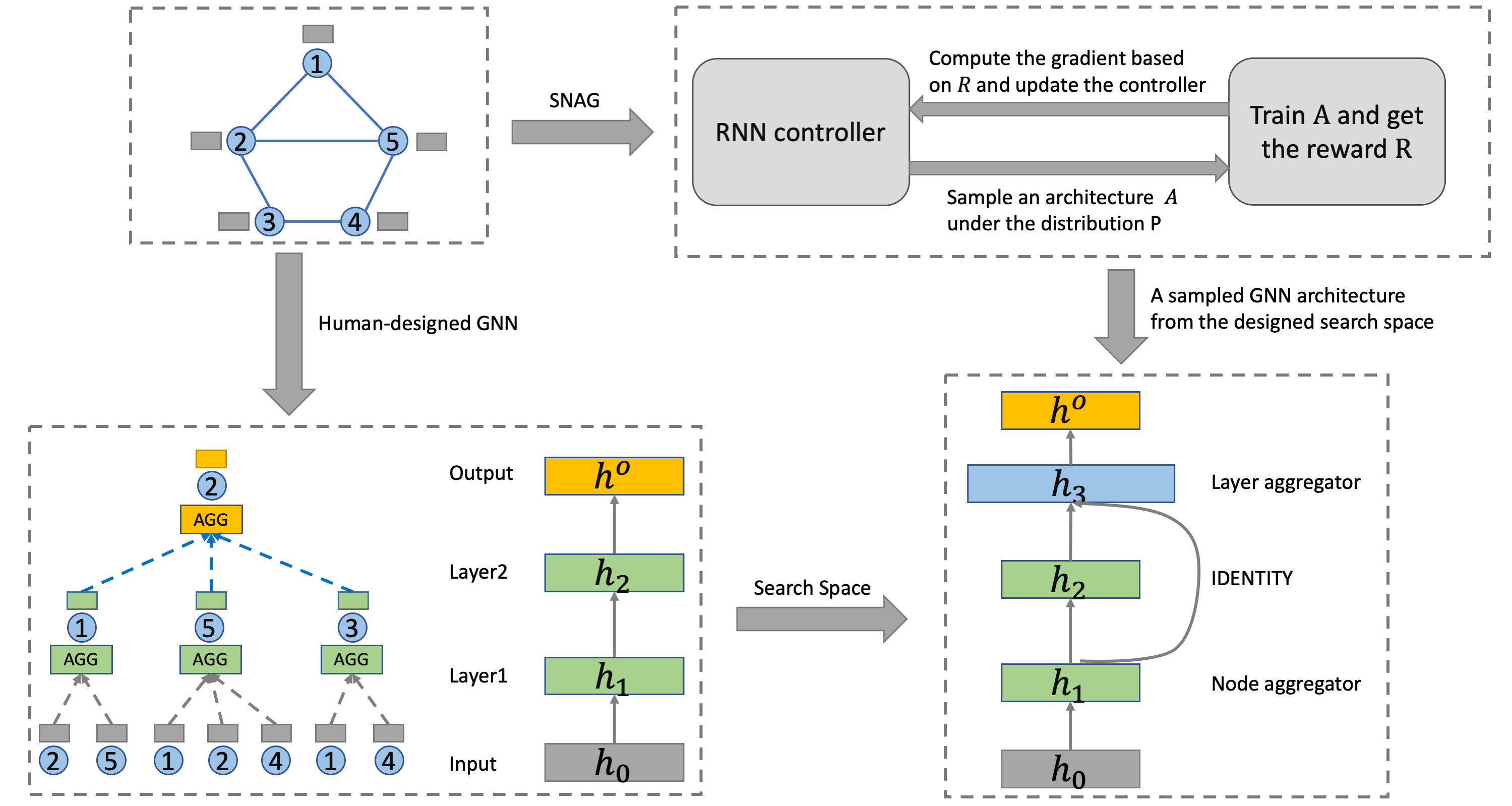}
	\caption{The whole framework of the propsed SNAG (Best view in color). a) Upper Left: an example graph with five nodes. The gray rectangle represent the input features attached to each node; b) Bottom Left: a typical 2-layer GNN architecture following the message passing neighborhood aggregation schema, which computes the embeddings of node ``2''; c) Upper Right: the reinforcement learning pipeline for NAS; d) Bottom Right: an illustration of a search space of the proposed SNAG using 2-layer GNN as backbone, which includes two key components of existing GNN models: node and layer aggregators.}
	\label{fig-framework}
	\vspace{0.1in}
\end{figure*}
\section{Introduction}
\label{sec-intro}
In recent years, Graph Neural Networks (GNN) \cite{gori2005new,battaglia2018relational} 
have been a hot topic due to their promising results on various graph-based tasks, e.g., recommendation~\cite{ying2018graph,wang2018billion,xiao2019beyond}, fraud detection \cite{liu2019geniepath}, chemistry \cite{gilmer2017neural}. 
In the literature, various GNN models \cite{kipf2016semi,hamilton2017inductive,velivckovic2017graph,gao2018large,xu2018powerful,xu2018representation,liu2019geniepath,xiao2019beyond} have been designed for graph-based tasks.
Despite the success of these GNN models, 
there are two challenges facing them. 
The first one is that there is no optimal architecture which can always behave well in different scenarios. 
For example, in our experiments (Table~\ref{tb-performance-trans} and \ref{tb-performance-induc}), 
we can see that the best GNN architectures vary on different datasets and tasks. It means that we have to spend huge computational and expertise resources designing and tuning a well-behaved GNN architecture given a specific task, which limits the application of GNN models. Secondly, existing GNN models do not make full use of the best architecture design practices in other established areas, e.g., computer vision (CV). For example, existing multi-layer GNN models tend to stack multiple layers with the same aggregator (see bottom left of Figure~\ref{fig-framework}), which aggregates hidden features of multi-hop neighbors. However it remains to be seen whether combinations of different aggregators in a multi-layer GNN model can further improve the performance. 
These challenges lead to a straightforward question: 
\textit{can we obtain well-behaved data-specific GNN architectures}?

To address the above question, researchers turn to neural architecture search (NAS)~\cite{baker2016designing,zoph2016neural,elsken2018neural,quanming2018taking} approaches, 
which have shown promising results in automatically designing architectures for convolutional neural networks (CNN) and recurrent neural networks (RNN). In very recent time, there are two preliminary works, GraphNAS~\cite{gao2019graphnas} and Auto-GNN~\cite{zhou2019auto}, making the first attempt to apply NAS to GNN architecture design.
Though GraphNAS and Auto-GNN show some promising results,  
there are some drawbacks in expressive capability and search efficiency of GraphNAS and Auto-GNN due to the designed search space. 
In the NAS literature~\cite{baker2016designing,zoph2016neural,elsken2018neural,quanming2018taking}, 
a good search space should be both expressive and compact. 
That is the search space should be large enough to subsume existing human-design architectures, 
thus the performance of a search method can be guaranteed. 
However, it will be extremely costly if the search space is too general, which is impractical for any searching method. 
The search spaces of GraphNAS and Auto-GNN are the same, both of which do not well satisfy the requirement of a good search space. 
On one hand, they fail to include several latest GNN models, e.g., the GeniePath~\cite{liu2019geniepath}, 
for which we give a more detailed analysis in Section~\ref{subsec-search-space} (Table~\ref{tb-search-space-comparison}). 
On the other hand, the search space includes too many choices, making it too complicated to search efficiently. 

In this work, to overcome the drawbacks of GraphNAS and Auto-GNN and push forward the research of NAS approaches for GNN, we propose the SNAG framework (Simplified Neural Architecture search for Graph neural networks), consisting of a simpler yet more expressive search space and a RL-based search algorithm. By revisiting extensive existing works, we unify state-of-the-art GNN models in a message passing framework~\cite{gilmer2017neural}, based on which a much more expressive yet simpler search space is designed. The simplified search space can not only emulate a series of existing GNN models, but also be very flexible to use the weight sharing mechanism, which is a widely used technique to accelerate the search algorithm in the NAS literature.
We conduct extensive experiments to demonstrate the effectiveness of the SNAG framework comparing to various baselines including GraphNAS and Auto-GNN.
To summarize, the contributions of this work are in the following: 
\begin{itemize}[leftmargin=10px, topsep = 2px]
\item In this work, to automatically obtain well-behaved data-specific GNN architectures, 
we propose the SNAG framework, 
which can
overcome the drawbacks of existing NAS approaches, i.e., GraphNAS and Auto-GNN. 
To better utilize the NAS techniques, 
we design a novel and effective search space, 
which can emulate more existing GNN architectures than previous works.

\item We design a 
RL-based search algorithm and its variant by adopting the weight sharing mechanism (SNAG-WS). 
By comparing the performance of these two variants, 
we show that the weight sharing mechanism is not empirically useful as we imagined, 
which aligns with the latest research in NAS literature~\cite{sciuto2020evaluating}. 

\item Extensive experiments on real-world datasets are conducted to evaluate the proposed SNAG framework, 
comparing to human-designed GNNs and NAS methods. 
The experimental results demonstrate the superiority of SNAG in terms of effectiveness and efficiency compared extensive baseline models. 
\end{itemize}


\begin{table*}[ht]
	\centering
	\caption{Comparisons of the search space between existing NAS methods and SNAG. For more details of the ``Others'' columns of GraphNAS/Auto-GNN, we refer readers to the corresponding papers.}
	\small
	\vspace{-10px}
	\label{tb-search-space-comparison}
	\begin{tabular}{C{50px} | C{150px} | c |C{110px}}
		\toprule
		& Node aggregators & Layer aggregators & Others \\\midrule
		GraphNAS/ Auto-GNN &   \texttt{GCN},\texttt{SAGE-SUM}/\texttt{-MEAN}/\texttt{-MAX}, \texttt{MLP}, \texttt{GAT} , \texttt{GAT-SYM}/\texttt{-COS}/ \texttt{-LINEAR}/\texttt{-GEN-LINEAR}  ,              &       -           &  \texttt{Hidden Embedding Size}, \texttt{Attention Head}, \texttt{Activation Function}  \\\hline
		Ours              &   All above plus \texttt{SAGE-LSTM} and \texttt{GeniePath}   & \texttt{CONCAT},\texttt{MAX},\texttt{LSTM}      &     \texttt{IDENTITY}, \texttt{ZERO} \\\bottomrule
	\end{tabular}
	\vspace{5px}
\end{table*}

\section{Related Works}
\label{sec-rel}

\subsection{Graph Neural Network (GNN)}
\label{sec-rel-gnn}

GNN is first proposed in~\cite{gori2005new}
and in the past five years many different variants~\cite{kipf2016semi,hamilton2017inductive,velivckovic2017graph,gao2018large,xu2018powerful,xu2018representation,liu2019geniepath} have been designed, all of which are relying on a neighborhood aggregation
(or \textit{message passing}) schema~\cite{gilmer2017neural}. As shown in the left part of 
Figure~\ref{fig-framework}, it tries to learn the representation of a given node in a graph by 
iteratively aggregating the hidden features (``message'') of its neighbors, and the message can propagate to farther neighborhood in the graph, e.g., the hidden features of two-hop neighbors can be aggregated in a two-step iteration process. 
Let $\mathcal{G} = (\mathcal{V}, \mathcal{E})$ be a simple graph with node features $\bX \in \mathbb{R}^{N \times d}$, 
where $\mathcal{V}$ and $\mathcal{E}$ represent the node and edge sets, respectively. $N$ represents the number of nodes and $d$ is the dimension of node features.
We use $N(v)$ to represent the first-order neighbors of a node $v$ in $\mathcal{G}$, i.e., $N(v) = \{u \in \mathcal{V}|(v,u) \in \mathcal{E}\}$. In the literature, we also create a new set $\widetilde{N}(v)$ is the neighbor set including itself, i.e.,  $\widetilde{N}(v) =  \{v\} \cup \{u \in \mathcal{V}|(v,u) \in \mathcal{E}\} $.


Then a $K$-layer GNN can be written as follows: the $l$-th layer
$(l = 1,\cdots, K)$ updates $\bh_v$ for each node $v$ by aggregating its neighborhood as
\begin{align}
	\label{eq-mpnn}
	\bh_v^{l} =  \sigma\bigg(\bW^{(l)} \cdot \Phi_n\bigg(\{\bh_u^{(l-1)}, \forall u \in \widetilde{N}(v)\}\bigg)\bigg),
\end{align}
where 
$\bh_v^{(l)} \in \mathbb{R}^{d_l}$ represents the hidden features of a node $v$ learned by the $l$-th layer, and $d_l$ is the corresponding dimension. $\bW^{(l)}$ is a trainable weight matrix shared by all nodes in the graph, and $\sigma$ is a non-linear activation function, e.g., a sigmoid or ReLU. $\Phi_n$ is the key component, i.e., a pre-defined aggregation function, which varies across on different GNN models. 
For example, in~\cite{kipf2016semi}, a weighted summation function is designed as the node aggregators, 
and in~\cite{hamilton2017inductive}, different functions, e.g., mean and max pooling, are proposed as the aggregators. 
Further, to weigh the importance of different neighbors, attention mechanism is incorporated to design the aggregators~\cite{velivckovic2017graph}. 

Usually, the output of the last layer is used as the final representation for each node, which is denoted as $\bz_v = \bh_v^{(K)}$. In~\cite{xu2018powerful}, skip-connections~\cite{he2016deep} are incorporated to propagate message from intermediate layers to an extra layer, and the final representation of the node $v$ is computed by a layer aggregation as $\bz_v  =  \Phi_l\left( \bh_v^{(1)},\cdots, \bh_v^{(K)} \right)$, and $\Phi_l$ can also have different options, e.g., max-pooling, concatenation. Based on the node and layer aggregators, we can define the two key components of exiting GNN models, i.e., the neighborhood aggregation function and the range of the neighborhood, which tends to be tuned depending on the tasks.
In Table~\ref{tb-search-space-comparison}, we list all node and layer aggregators in this work, which lays the basis for the proposed SNAG framework.

\subsection{Neural Architecture Search (NAS)}
\label{sec:nas}

Neural architecture search (NAS)~\cite{baker2016designing,zoph2016neural,elsken2018neural,quanming2018taking} 
aims to automatically find better and smaller architectures comparing to expert-designed ones, which have shown promising results in architecture design for CNN and Recurrent Neural Network (RNN)~\cite{liu2018darts,zoph2018learning,tan2019efficientnet,yao2020efficient,zhang2019neural}. 
In the literature, one of the representative NAS approaches are reinforcement learning (RL)~\cite{baker2016designing,zoph2016neural,pham2018efficient}, which trains an RNN controller in the loop: the controller firstly generates an candidate architecture by sampling a list of actions (operations) from a pre-defined search space, and then trains it to convergence to obtain the performance of the given task. The controller then uses the performance as the guiding signal to update the RNN parameters, and the whole process is repeated for many iterations to find more promising architectures. GraphNAS~\cite{gao2019graphnas} and Auto-GNN~\cite{zhou2019auto} are the first two RL-based NAS methods for GNN.

Search space is a key component of NAS approaches, the quality of which directly affects the final performance and search efficiency. 
As mentioned in \cite{baker2016designing,zoph2016neural,zoph2018learning,Bender18one-shot,pham2018efficient,liu2018darts,li2019random,zhang2019neural,yao2020efficient}, a good search space should include existing human-designed models, thus the performance of an designed search algorithm can be guaranteed. 
In this work, by unifying existing GNN models in the message passing framework~\cite{gilmer2017neural} with the proposed node and layer aggregators, we design a more expressive yet simple search space in this work, which is also flexible enough to incorporate the weight sharing mechanism into our RL-based method.

\section{The Proposed Framework}
\label{sec-framework}

\subsection{The design of search space}
\label{subsec-search-space}

As introduced in Section~\ref{sec-rel-gnn}, 
most existing GNN architectures are relying on a message passing framework~\cite{gilmer2017neural}, 
which constitutes the backbone of the designed search space in this work. 
Besides, motivated by JK-Network~\cite{xu2018representation}, 
to further improve the expressive capability, 
we modify the message framework by adding an extra layer  which can adaptively combine the outputs of all node aggregation layers. 
In this work, we argue and demonstrate in the experiments that these two components are the key parts for a well-behaved GNN model, 
denoted as \textit{Node arggregators} and \textit{Layer aggregators}. The former one focus on how to aggregate the neighborhood features, 
while the latter one focus on the range of neighborhood to use.  Here we introduce the backbone of the proposed search space, 
as shown in the bottom right part of Figure~\ref{fig-framework}, which consists of two key components:
\begin{itemize}[leftmargin=10px]
	\item \textbf{Node aggregators}: 
	We choose 12 node aggregators based on popular GNN models, and they are presented in 
	Table~\ref{tb-search-space-comparison}. 
	
	\item \textbf{Layer aggregators}: 
	We choose 3 layer aggregators as shown in Table~\ref{tb-search-space-comparison}. Besides, we have two more operations, \texttt{IDENTITY} and \texttt{ZERO}, related to skip-connections. Instead of requiring skip-connections between all intermediate layers and the final layer in JK-Network, in this work, we generalize this option by proposing to search for the existence of skip-connection between each intermediate layer and the last layer. To connect, we choose \texttt{IDENTITY}, and \texttt{ZERO} otherwise. 
\end{itemize}

To further inject the domain knowledge from existing GNN architectures, when searching for the skip-connections for each GNN layer, we add one more constraint that the last layer should always be used as the final output, thus for a $K$-layer GNN architecture, we need to search $K-1$ \texttt{IDENTITY} or \texttt{ZERO} for the skip-connection options.

\subsection{Problem formulation}
After designing the search space, denoted as $\mathcal{A}$, the search process implies a bi-level optimization problem~\cite{colson2007overview,franceschi2018bilevel}, as show in the following:
\begin{align}
\min\nolimits_{\alpha \in \mathcal{A}} 
& \quad\mathcal{L}_{val} (\alpha, w^*),
\label{eq-nas-opt}
\\
\text{s.t.} 
& \quad w^* = \arg\min\nolimits_w \mathcal{L}_{train} (\alpha, w),
\notag
\end{align}
where $\mathcal{L}_{train}$ and $\mathcal{L}_{val}$ represent the training and validation loss, respectively, and $\mathcal{A}$ represents the search space introduced in Section~\ref{subsec-search-space}. $\alpha$ and $w$ represent the architecture and model parameters. Eq.~\eqref{eq-nas-opt} denotes a trial-and-error process for the NAS problem, which selects an architecture $\alpha$ from the search space, and then trains it from scratch to obtain the best performance. This process is repeated during the given time budget and the optimal $\alpha^*$ is kept track of and returned after the search process finished.

In this work, motivated by the pioneering NAS works~\cite{baker2016designing,zoph2016neural}, 
we design a RL method to execute the search process. 
To be specific, during the search phase, 
we use a recurrent neural network (RNN) controller, 
parameterized by $\theta_c$, to sample an candidate architecture from the search space. 
The architecture is represented by a list of actions (OPs), including the node aggregators, 
layer aggregators and \texttt{IDENTITY}/\texttt{ZERO} as shown in Table~\ref{tb-search-space-comparison}. 
Then the candidate architecture will be trained till convergence, 
and the accuracy on a held-out validation set $\mathcal{D}_{val}$ is returned. 
The parameters of the RNN controller are then optimized in order to maximize the expected validation accuracy $\mathbb{E}_{P(\alpha; \theta_c)}[\mathcal{R}]$ on $\mathcal{D}_{val}$, 
where $P(\alpha;\theta_c)$ is the distribution of architectures parameterized by $\theta_c$, and $\mathcal{R}$ is the validation accuracy. 
In this way, the RNN controller will generate better architectures over time, and can obtain optimal one in the end of the search phase. 
After finishing the search process, 
we need to derive the searched architectures. We first sample $n$ architectures under the trained distribution $P(\alpha, \theta_c)$, and for each architecture, we train them from scratch with some hyper-parameters tuning, e.g., the embedding size and learning rate, etc. We then select the best architecture as the searched one, which aligns with the process in previous works~\cite{zoph2016neural,pham2018efficient}. In our experiments, we empirically set $n =10$ for simplicity. For more technical details, we refer readers to~\cite{zoph2016neural,gao2019graphnas}.

Besides, in this work, we also incorporate the weight sharing mechanism into our framework, and propose the 
SNAG-WS variant. The key difference between SNAG and SNAG-WS lies in that we create a dictionary to load and save the trained parameters of all OPs (Table~\ref{tb-search-space-comparison}) in a sampled architecture during the search process.

\begin{table}[ht]
	\vspace{5px}
	\caption{Dataset statistics of the datasets in the experiments.}
	\label{tb-datasets}
	\centering
	\vspace{-10px}
	\begin{tabular}{l|ccc|c}
		\toprule
		& \multicolumn{3}{c|}{Transductive}                                         & Inductive \\ \cline{2-5} 
		& Cora  & CiteSeer & PubMed & PPI       \\ \midrule
		\#nodes    & 2,708 & 3,327    & 19,717 & 56,944    \\ \hline
		\#edges    & 5,278 & 4,552    & 44,324 & 818,716   \\ \hline
		\#features & 1,433 & 3,703    & 500    & 121       \\ \hline
		\#classes  & 7     & 6        & 3      & 50        \\ \bottomrule
	\end{tabular}
\end{table}

\begin{table*}[ht]
	\caption{Performance comparisons in transductive tasks. We show the mean classification accuracy (with standard deviation). We categorize baselines into human-designed GNNs and NAS methods. The best results in different groups of baselines are underlined, and the best result on each dataset is in boldface.}
	\small
	\centering
	\vspace{-10px}
	\begin{tabular}{C{48px}|L{68px}|ccc}
		\toprule
		&              & \multicolumn{3}{c}{Transductive}        \\\cline{2-5}
		&    Methods           & Cora            & CiteSeer        & PubMed      \\\midrule
		\multirow{11}{28px}{Human-designed GNN}
		& GCN    & 0.8761 (0.0101) & 0.7666 (0.0202) & 0.8708 (0.0030) \\
		& GCN-JK        & 0.8770 (0.0118) & \underline{0.7713 (0.0136)} & 0.8777 (0.0037) \\
		& GraphSAGE          & 0.8741 (0.0159) & 0.7599 (0.0094) & 0.8784 (0.0044)\\
		& GraphSAGE-JK       & \underline{0.8841 (0.0015)} & 0.7654 (0.0054) & {0.8822 (0.0066)}\\
		& GAT           & 0.8719 (0.0163) & 0.7518 (0.0145) & 0.8573 (0.0066) \\
		& GAT-JK        & 0.8726 (0.0086) & 0.7527 (0.0128) & 0.8674 (0.0055) \\
		& GIN           & 0.8600 (0.0083) & 0.7340 (0.0139) & 0.8799 (0.0046) \\
		& GIN-JK        & 0.8699 (0.0103) & 0.7651 (0.0133) & \underline{0.8828 (0.0054)} \\
		& GeniePath     & 0.8670 (0.0123) & 0.7594 (0.0137) & 0.8796 (0.0039) \\
		& GeniePath-JK  & 0.8776 (0.0117)  & 0.7591 (0.0116) & 0.8818 (0.0037) \\
		& LGCN          & 0.8687 (0.0075) &       0.7543 (0.0221)           &    0.8753 (0.0012)       \\\midrule
		\multirow{5}{28px}{NAS methods}  & Random & 0.8694 (0.0032) & \textbf{\underline{0.7820 (0.0020)}} & 0.8888(0.0009)\\
		& Bayesian      & 0.8580 (0.0027) & 0.7650 (0.0021) & 0.8842(0.0005)  \\
		& GraphNAS      & \underline{0.8840 (0.0071)}  & 0.7762 (0.0061) &      { \underline{0.8896 (0.0024)}}         \\
		& GraphNAS-WS &     0.8808 (0.0101)            &    0.7613 (0.0156)             & 0.8842 (0.0103)     \\\midrule
		\multirow{2}{*}{ours} 
		& SNAG   &0.8826 (0.0023)  &0.7707 (0.0064) &0.8877 (0.0012) \\
		& SNAG-WS & \textbf{0.8895 (0.0051)} &0.7695 (0.0069) &\textbf{0.8942 (0.0010)}\\
		\bottomrule
	\end{tabular}
	\label{tb-performance-trans}
	\vspace{5px}
\end{table*}

\begin{table}[ht]
	\caption{Performance comparisons in inductive tasks. We show the Micro-F1 (with standard deviation). We categorize baselines into human-designed GNNs and NAS methods. The best results in different groups of baselines are underlined, and the best result is in boldface.}
	\small
	\centering
	\vspace{-10px}
	\begin{tabular}{c|l|c }
		\toprule
		&    Methods           & PPI   \\\midrule
		\multirow{10}{65px}{Human-designed GNN}
		& GCN    &0.9333 (0.0019) \\
		& GCN-JK        &0.9344 (0.0007) \\
		& GraphSAGE      &0.9721 (0.0010)\\
		& GraphSAGE-JK       &0.9718 (0.0014)  \\
		& GAT          &\underline{0.9782 (0.0005)} \\
		& GAT-JK        &0.9781 (0.0003) \\
		& GIN           &0.9593 (0.0052)  \\
		& GIN-JK        &0.9641 (0.0029) \\
		& GeniePath           &0.9528 (0.0000)  \\
		& GeniePath-JK           &0.9644 (0.0000)  \\\midrule
		\multirow{5}{*}{NAS methods}  & Random  &0.9882 (0.0011)     \\
		& Bayesian      &  \textbf{\underline{0.9897 (0.0008)}}\\
		& GraphNAS      & 0.9698 (0.0128)    \\
		& GraphNAS-WS &  0.9584 (0.0415)\\\midrule
		\multirow{2}{*}{ours} & SNAG          &\textbf{0.9887 (0.0010)}  \\
		& SNAG-WS       &0.9875 (0.0006) \\
		\bottomrule
	\end{tabular}
	\label{tb-performance-induc}
\end{table}
\section{Experiments}
\label{sec-exp}


\subsection{Experimental Settings}
\subsubsection{Datasets and Tasks.}
Here, we introduce two tasks and the corresponding datasets (Table~\ref{tb-datasets}), which are standard ones in the literature~\cite{kipf2016semi,hamilton2017inductive,xu2018representation}.

\noindent \textbf{Transductive Task.} 
Only a subset of nodes in one graph are used as training data, and other nodes are used as validation and test data. 
For this setting, we use three benchmark dataset: Cora, CiteSeer, PubMed. 
They are all citation networks, provided by~\cite{sen2008collective}. 
Each node represents a paper, and each edge represents the citation relation between two papers. 
The datasets contain bag-of-words features for each paper (node), and the task is to classify papers into different subjects based on the citation networks.

For all datasets, We split the nodes in all graphs into 60\%, 20\%, 20\% for training, validation, and test. For the transductive task, we use the classification accuracy as the evaluation metric.

\noindent \textbf{Inductive Task.} 
In this task, we use a number of graphs as training data, and other completely unseen graphs as validation/test data. 
For this setting, we use the PPI dataset, provided by~\cite{hamilton2017inductive}, on which the task is to classify protein functions. 
PPI consists of 24 graphs, with each corresponding to a human tissue. Each node has positional gene sets, motif gene sets and immunological signatures as features and gene ontology sets as labels. 20 graphs are used for training, 2 graphs are used for validation and the rest for testing, respectively. For the inductive task, we use Micro-F1 as the evaluation metric.

\subsubsection{Compared Methods}
We compare SNAG with two groups of state-of-the-art methods: human-designed GNN architectures and NAS methods for GNN.

\noindent
\textbf{Human-designed GNNs.} We use the following popular GNN architectures: GCN \cite{kipf2016semi}, GraphSAGE \cite{hamilton2017inductive}, GAT \cite{velivckovic2017graph}, GIN \cite{xu2018powerful}, LGCN \cite{gao2018large}, GeniePath \cite{liu2019geniepath}. For models with variants, like different aggregators in GraphSAGE or different attention functions in GAT, we report the best performance across the variants. Besides, we extend the idea of JK-Network~\cite{xu2018representation} in all models except for LGCN, and obtain 5 more baselines: GCN-JK, GraphSAGE-JK, GAT-JK, GIN-JK, GeniePath-JK, which add an extra layer. For LGCN, we use the code released by the authors~\footnote{https://github.com/HongyangGao/LGCN}. For other baselines, we use the popular open source library Pytorch Geometric (PyG)~\cite{Fey/Lenssen/2019}~\footnote{https://github.com/rusty1s/pytorch\_geometric}, which implements various GNN models. 
For all baselines, we train it from scratch with the obtained best hyper-parameters on validation datasets, and get the test performance. We repeat this process for 5 times, and report the final mean accuracy with standard deviation.

\noindent
\textbf{NAS methods for GNN.}
We consider the following methods: Random search (denoted as ``Random'') and Bayesian optimization \cite{bergstra2011algorithms} (denoted as ``Bayesian''), which directly search on the search with random sampling and bayesian optimization methods, respectively. Besides, GraphNAS\footnote{\url{https://github.com/GraphNAS/GraphNAS}}~\cite{gao2019graphnas} is chosen as NAS baseline.

Note that for human-designed GNNs and NAS methods, for fair comparison and good balance between efficiency and performance, we choose set the number of GNN layers to be 3, which is an empirically good choice in the literature~\cite{velivckovic2017graph,liu2019geniepath}.



\begin{figure*}[ht]
	\subfigure[Cora.]{\includegraphics[width=0.24\textwidth]{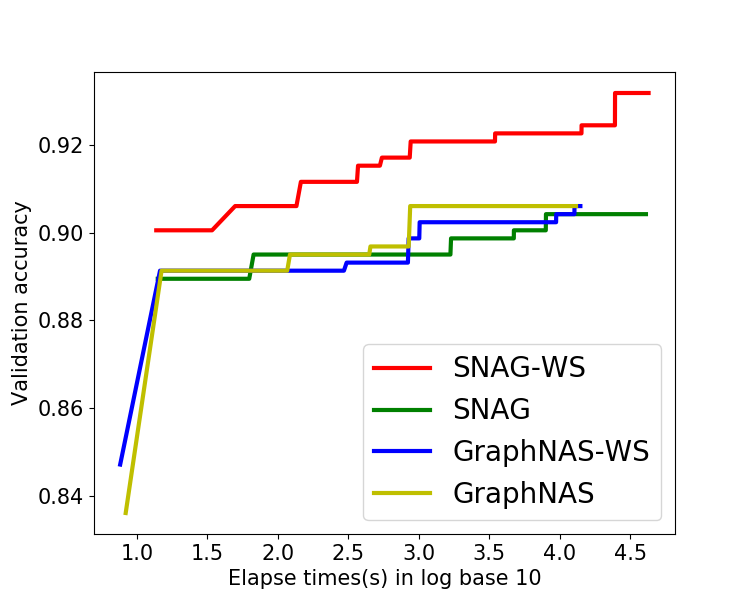}}
	\subfigure[CiteSeer.]{\includegraphics[width=0.24\textwidth]{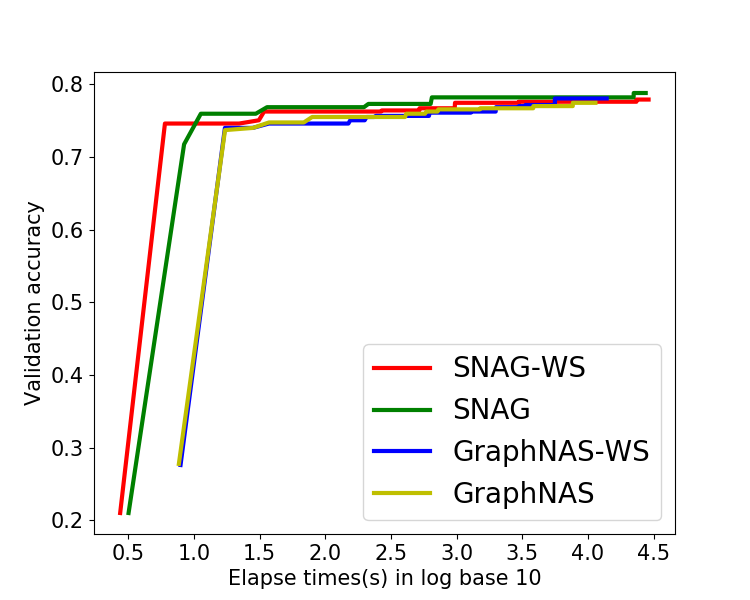}}
	\subfigure[PubMed.]{\includegraphics[width=0.24\textwidth]{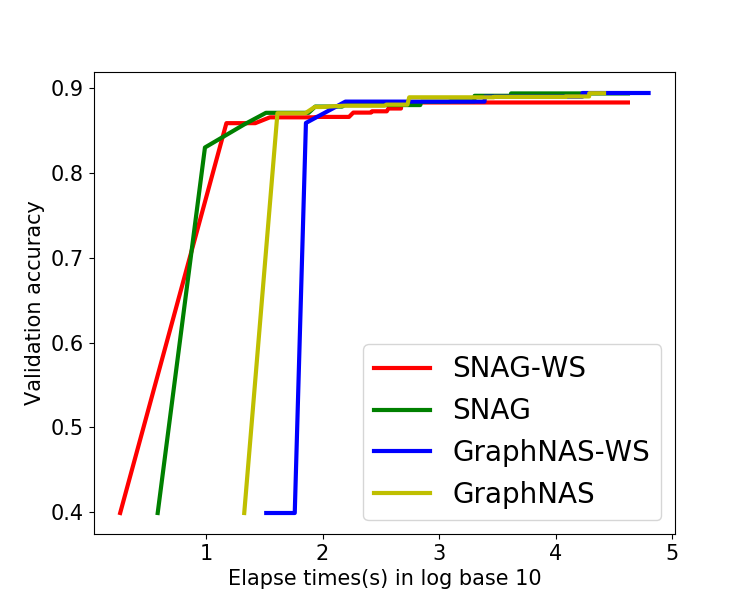}}
	\subfigure[PPI.]{\includegraphics[width=0.24\textwidth]{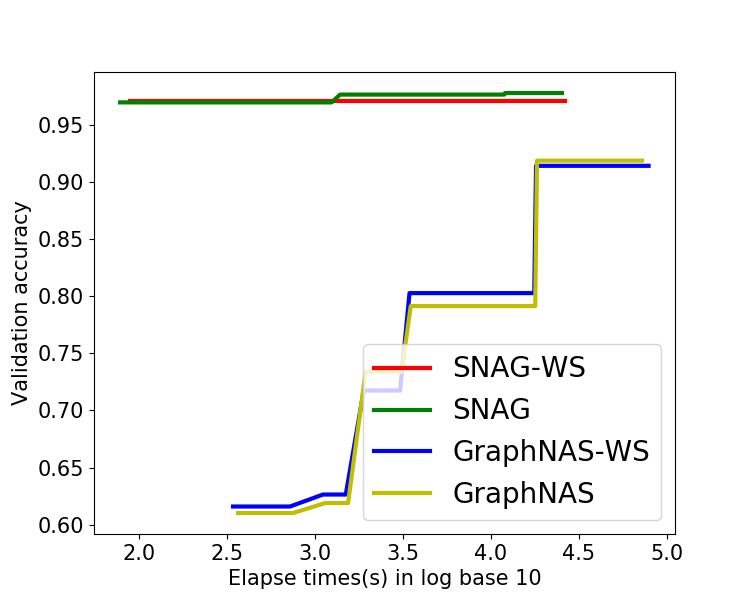}}
	\centering
	\vspace{-10px}
	\caption{The validation accuracy w.r.t. search time (in seconds) in log base 10.}
	\label{fig-search-efficiency}
\end{figure*}

\begin{table*}[ht]
	\caption{Performance comparisons of SNAG and SNAG-WS using different search spaces.}
	\centering\small
	\vspace{-10px}
	\begin{tabular}{c|c|c|c|c}
		\toprule
		& \multicolumn{2}{c|}{SNAG} &\multicolumn{2}{c}{SNAG-WS}   \\\cline{2-5}		
		& layer aggregators (w)  &layer aggregators (w/o) &layer aggregators (w)  &layer aggregators (w/o)    \\\midrule		
		Cora &0.8826 (0.0023)&0.8822 (0.0071) &0.8895 (0.0051)&0.8892 (0.0062)\\
		CiteSeer &0.7707 (0.0064)& 0.7335 (0.0025)&0.7695 (0.0069)&0.7530 (00034)\\
		PubMed & 0.8877 (0.0012)& 0.8756 (0.0016)&0.8942 (0.0010) &0.8800 (0.0013)\\
		PPI &0.9887 (0.0010)&0.9849 (0.0040)&0.9875 (0.0006)& 0.9861 (0.0009)\\	
		\bottomrule
	\end{tabular}
	\label{tb-performance-jk}
	\vspace{5px}
\end{table*}

%

\subsection{Performance comparison}
\label{subsec-performance-comparison}
In this part, we give the analysis of the performance comparisons on different datasets. 



From Table~\ref{tb-performance-trans}, 
we can see that SNAG models, including SNAG-WS, win over all baselines on most datasets except CiteSeer. Considering the fact that the performance of SNAG on CiteSeer is very close to the best one (Random), it demonstrates the effectiveness of the NAS methods on GNN. In other words, with SNAG, we can obtain well-behaved GNN architectures given a new task. When comparing SNAG with GraphNAS methods, the performance gain is evident. We attribute this to the superiority of the expressive yet simple search space. 


From Table~\ref{tb-performance-induc}, 
we can see that the performance trending is very similar to that in transductive task, 
which is that the NAS methods can obtain better or competitive performance than human-designed GNNs. 
When looking at the NAS methods, we can see that our proposed SNAG, 
Random and Bayesian outperforms GraphNAS. 
This also demonstrates the superiority of the designed search space.

Taking into consideration the results of these two tasks, we demonstrate the effectiveness of SNAG models, especially the superiority of the search space. 



\subsection{Understanding the search space of SNAG}
\label{subsec-understand-serach-space}
In this section, we show the simplicity and expressiveness of the designed search space of SNAG from two aspects: speedup in searching and the performance gain from the layer aggregators.

\subsubsection{Speedup in searching}
In this part, to show the simplicity of the designed search space, we compare the efficiency of SNAG and GraphNAS by showing the validation accuracy w.r.t to the running time, and the results are shown in Figure~\ref{fig-search-efficiency}. The accuracy is obtained by evaluating the sampled architecture on validation set after training it from scratch till convergency, which can reflect the capability of NAS methods in discovering better architectures with time elapsing. From Figure~\ref{fig-search-efficiency}, we can see that SNAG speeds up the search process significantly comparing to GraphNAS, i.e., the model can obtain better GNN architectures during the search space. Considering the fact that both GraphNAS and SNAG adopt the same RL framework, then this advantage is attributed to simpler and smaller search space.

\subsubsection{Influence of layer aggregators}
In this part, to show the stronger expressive capability of the designed search space, we conduct experiments on all datasets using a search space only with the node aggregators, i.e., removing the layer aggregators, as comparisons. The results are shown in Table~\ref{tb-performance-jk}, and we report the test accuracies of both the SNAG and SNAG-WS. From Table~\ref{tb-performance-jk}, we can see that the performance consistently drops on all datasets when removing the layer aggregators, which demonstrates the importance of the layer aggregators for the final performance and aligns with the observation in~Section~\ref{subsec-performance-comparison} that the performance of human-designed GNNs can be improved by adopting the JK-Network architecture.

\section{Conclusion and Future work}
\label{sec-conclusion}
In this work, to overcome the drawbacks in expressive capability and search efficiency of two existing NAS approaches for GNN, 
i.e., GraphNAS~\cite{gao2019graphnas} and Auto-GNN~\cite{zhou2019auto}, we propose the SNAG framework, 
i.e., Simplified Neural Architecture search for GNN. 
By revisiting existing works, we unify state-of-the-art GNN models in a message passing framework~\cite{gilmer2017neural}, 
and design a simpler yet more expressive search space than that of GraphNAS and Auto-GNN. A RL-based search algorithm is designed and a variant (SNAG-WS) is also proposed by incorporating the weight sharing mechanism. Through extensive experiments on real-world datasets, we not only demonstrate the effectiveness of the proposed SNAG framework comparing to various baselines including GraphNAS and Auto-GNN, but also give better understanding of different components of the proposed SNAG. 
For future work, we will explore the SNAG framework in more graph-based tasks besides node classification.

\bibliographystyle{plain}
\bibliography{main-ws}

\end{document}